\documentclass[11pt,a4paper]{article}
\usepackage[hyperref]{acl2019}
\usepackage{times}
\usepackage{latexsym}
\usepackage{graphicx}

\usepackage{url}

\aclfinalcopy 
\def\aclpaperid{167} 

\title{Scheduled Sampling for Transformers}

\author{
  Tsvetomila Mihaylova \\
  Instituto de Telecomunica\c{c}\~oes \\
  Lisbon, Portugal \\
  \texttt{tsvetomila.mihaylova@lx.it.pt}
  \And
  Andr\'{e} F.~T. Martins \\
  Instituto de Telecomunica\c{c}\~oes \& Unbabel \\
  Lisbon, Portugal\\
  \texttt{andre.martins@unbabel.com}
  }

\date{}

\begin{document}
\maketitle
\begin{abstract}
  Scheduled sampling is a technique for avoiding one of the known problems in sequence-to-sequence generation: exposure bias. It consists of feeding the model a mix of the teacher forced embeddings and the model predictions from the previous step in training time. The technique has been used for improving the model performance with recurrent neural networks (RNN). In the Transformer model, unlike the RNN, the generation of a new word attends to the full sentence generated so far, not only to the last word, and it is not straightforward to apply the scheduled sampling technique.
  We propose some structural changes to allow scheduled sampling to be applied to Transformer architecture, via a two-pass decoding strategy. Experiments on two language pairs achieve performance close to a teacher-forcing baseline and show that this technique is promising for further exploration. 
\end{abstract}

\section{Introduction}

Recent work in Neural Machine Translation (NMT) relies on a sequence-to-sequence model with global attention \citep{sutskever2014sequence,bahdanau2014neural}, trained with maximum likelihood estimation (MLE). These models are typically trained by teacher forcing, in which the model makes each decision conditioned on the gold history of the target sequence. This tends to lead to quick convergence but is dissimilar to the procedure used at decoding time, when the gold target sequence is not available and decisions are conditioned on previous model predictions.


\citet{ranzato2015sequence}  point out the problem that using teacher forcing means the model has never been trained on its own errors and may not be robust to them---a phenomenon called \emph{exposure bias}. This has the potential to cause problems at translation time, when the model is exposed to its own (likely imperfect) predictions. 

A common approach for addressing the problem with exposure bias is using a scheduled strategy for deciding when to use teacher forcing and when not to \citep{bengio2015scheduled}. For a recurrent decoder, applying scheduled sampling is trivial: for generation of each word, the model decides whether to condition on the gold embedding from the given target (teacher forcing) or the model prediction from the previous step. 

In the Transformer model \citep{vaswani2017attention}, the decoding is still autoregressive, but unlike the RNN decoder, the generation of each word conditions on the whole prefix sequence and not only on the last word. This makes it non-trivial to apply scheduled sampling directly for this model. Since the Transformer achieves state-of-the-art results and has become a default choice for many natural language processing problems, it is interesting to adapt and explore the idea of scheduled sampling for it, and, to our knowledge, no way of doing this has been proposed so far.


Our contributions in this paper are:

\begin{itemize}
    \item We propose a new strategy for using scheduled sampling in Transformer models by making two passes through the decoder in training time.
    \item We compare several approaches for conditioning on the model predictions when they are used instead of the gold target.
    \item We test the scheduled sampling with transformers in a machine translation task on two language pairs and achieve results close to a teacher forcing baseline (with a slight improvement of up to 1 BLEU point for some models).
\end{itemize}

\section{Related Work}

\citet{bengio2015scheduled} proposed  scheduled sampling for sequence-to-sequence RNN models: a method where the embedding used as the input to the decoder at time step $t+1$ is picked randomly between the gold target and the \texttt{argmax} of the model's output probabilities at step $t$. The Bernoulli probability of picking one or the other changes over training epochs according to a schedule that makes the probability of choosing the gold target decrease across training steps. The authors propose three different schedules: linear decay, exponential decay and inverse sigmoid decay.

\citet{goyal2017differentiable} proposed an approach based on scheduled sampling which backpropagates through the model decisions. At each step, when the model decides to use model predictions, instead of the \texttt{argmax}, they use a weighted average of all word embeddings, weighted by the prediction scores. They experimented with two options: a softmax with a temperature parameter, and a stochastic variant using Gumbel Softmax \citep{jang2016categorical} with temperature. With this technique, they achieve better results than the standard scheduled sampling. 
Our works extends \citet{bengio2015scheduled} and \citet{goyal2017differentiable} by adapting their frameworks to Transformer architectures.


\citet{ranzato2015sequence} took ideas from scheduled sampling and the REINFORCE  algorithm \citep{williams1992simple} and combine the teacher forcing training with optimization of the sequence level loss. In the first epochs, the model is trained with teacher forcing and for the remaining epochs they start with teacher forcing for the first $t$ time steps and use REINFORCE (sampling from the model) until the end of the sequence. They decrease the time for training with teacher forcing $t$ as training continues until the whole sequence is trained with REINFORCE in the final epochs.  
In addition to the work of \citet{ranzato2015sequence} other methods that are also focused on sequence-level training are using for example  actor-critic \citep{bahdanau2016actor} or beam search optimization \citep{wiseman2016sequence}. These methods directly optimize the metric used at test time (e.g. BLEU). Another proposed approach to avoid exposure bias is SEARN \citep{daume2009search}. In SEARN, the model uses its own predictions at training time to produce sequence of actions, then a search algorithm determines the optimal action at each step and a policy is trained to predict that action. 
The main drawback of these approaches is that the training becomes much slower. By contrast, in this paper we focus on methods which are comparable in training time with a force-decoding baseline.

\section{Scheduled Sampling with Transformers}

In the case with recurrent neural networks (RNN) in the training phase we generate one word at a time step, and we condition the generation of this word to the previous word from the gold target sequence. This sequential decoding makes it simple to apply scheduled sampling - at each time step, with some probability, instead of using the previous word in the gold sequence, we use the word predicted from the model on the previous step.

The Transformer model \citep{vaswani2017attention}, which achieves state-of-the-art results for a lot of natural language processing tasks, is also an autoregressive model. The generation of each word conditions on all previous words in the sequence, not only on the last generated word. The model is based on several \textit{self-attention layers}, 
which directly model relationships between all words in the sentence, regardless of their respective position. The order of the words is achieved by position embeddings which are summed with the corresponding word embeddings. Using position masking in the decoder ensures that the generation of each word depends only on the previous words in the sequence and not on the following ones. 
Because generation of a word in the Transformer conditions on all previous words in the sequence and not just the last word, it is not trivial to apply scheduled sampling to it, where, in training time, we need to choose between using the gold target word or the model prediction. 
In order to allow usage of scheduled sampling with the Transformer model, we needed to make some changes in the Transformer architecture. 




\subsection{Two-decoder Transformer}

\begin{figure*}[h!]
    \centering
    \includegraphics[width=400px]{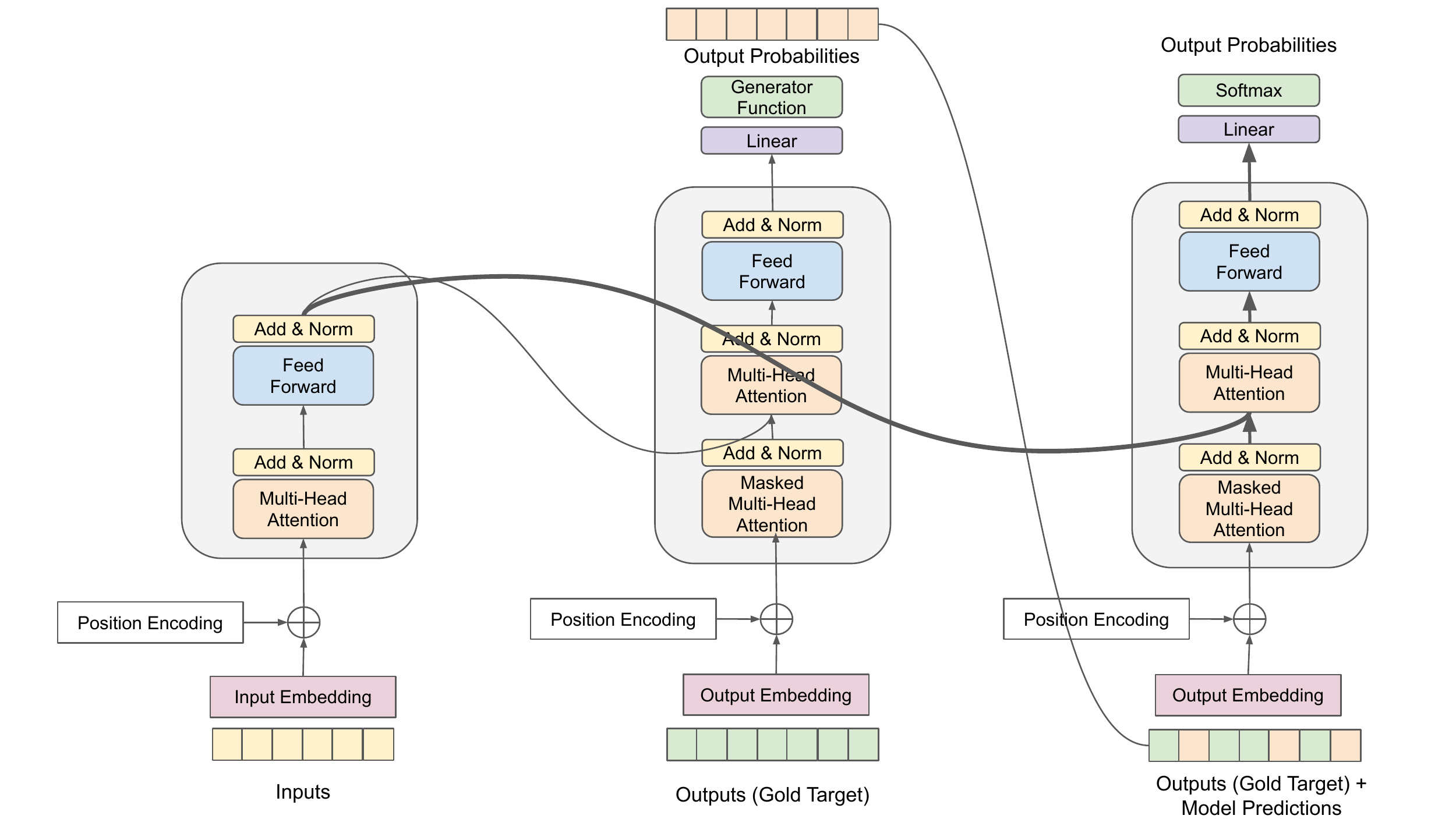}
    \caption{Transformer model adapted for use with scheduled sampling. The two decoders on the image share the same parameters. The first pass on the decoder conditions on the gold target sequence and returns the model predictions. The second pass conditions on a mix of the target sequence and model predictions and returns the result. The thicker lines show the path that is backpropagated in all experiments, i.e. we always make backpropagation through the second decoder pass. The thin arrows are only backpropagated in a part of the experiments. (\textit{The image is based on the transformer architecture from the paper of \citet{vaswani2017attention}.})}
    \label{figure:scheduled_transformer}
\end{figure*}

The model we propose for applying scheduled sampling in transformers makes two passes on the decoder. Its architecture is illustrated on Figure \ref{figure:scheduled_transformer}. We make no changes in the encoder of the model. The decoding of the scheduled transformer has the following steps:

\begin{enumerate}
    \item \textbf{First pass on the decoder: get the model predictions.} On this step, the decoder conditions on the gold target sequence and predicts scores for each position as a standard transformer model. Those scores are passed to the next step. 
    
    \item \textbf{Mix the gold target sequence with the predicted  sequence.}
    After obtaining a sequence representing the prediction from the model for each position, we imitate scheduled sampling by mixing the target sequence with the model predictions: 
    For each position in the sequence, we select with a given probability whether to use the gold token or the prediction from the model. The probability for using teacher forcing (i.e. the gold token) is a function of the training step and is calculated with a selected schedule.
    We pass this ``new reference sequence'' as the reference for the second decoder.
    The vectors used from the model predictions can be either the embedding of the highest-scored word, or a mix of the embeddings according to their scores. Several variants of building the vector from the model predictions for each position are described below.
    
    
    \item \textbf{Second pass on the decoder: the final predictions.}
    The second pass of the decoder uses as output target the mix of words in the gold sequence and the model predictions. The outputs of this decoder pass are the actual result from the models. 
    
\end{enumerate}

It is important to mention that the two decoders are identical and share the same parameters. We are using the same decoder for the first pass, where we condition on the gold sequence and the second pass, where we condition on the mix between the gold sequence and the model predictions.


\subsection{Embedding Mix}

For each position in the sequence, the first decoder pass gives a score for each vocabulary word. 
We explore several ways of using those scores when the model predictions are used.


\begin{itemize}
    \item The most obvious case is to not mix the embeddings at all and pass the \texttt{argmax} from the model predictions, i.e. use the embedding of the vocabulary word with the highest score from the decoder.
    
    \item We also experiment with mixing the \texttt{top-k} embeddings. In our experiments, we use the weighted average of the embeddings of the top-5 scored vocabulary words.
    
    \item Inspired by the work of \citet{goyal2017differentiable}, we experiment with passing a mix of the embeddings with \texttt{softmax} with temperature. Using a higher temperature parameter makes a better approximation of the \texttt{argmax}.
    
    \[\bar{e}_{i-1} = \sum_{y} e(y) \frac{\exp(\alpha s_{i-1}(y))}{\sum_{y'} \exp(\alpha s_{i-1}(y'))}\]
    where $\bar{e}_{i-1}$ is the vector which will be used at the current position, obtained by a sum of the embeddings of all vocabulary words, weighted by a softmax of the scores $s_{i-1}$.
    
    \item An alternative of using argmax is sampling an embedding from the softmax distribution. Also based on the work of \citet{goyal2017differentiable}, we use the Gumbel Softmax \citep{maddison2016concrete,jang2016categorical} approximation to sample the embedding: 
    \[\bar{e}_{i-1} = \sum_{y} e(y) \frac{\exp(\alpha( s_{i-1}(y)) + G_{y})}{\sum_{y'} \exp(\alpha(s_{i-1}(y') + G_{y'}))}\]
    
    where $U\sim \mathrm{Uniform}(0, 1)$ and $G = -\log(-\log U)$. 
    
    \item Finally, we experiment with passing a \texttt{sparsemax} mix of the embeddings \citep{martins2016softmax}.
\end{itemize}



\subsection{Weights update}

We calculate Cross Entropy Loss based on the outputs from the second decoder pass. 
For the cases where all vocabulary words are summed (Softmax, Gumbel softmax, Sparsemax), we try two variants of updating the model weights. 
\begin{itemize}
    \item Only backpropagate through the decoder which makes the final predictions, based on mix between the gold target and the model predictions.
    \item Backpropagate through the second, as well as through the first decoder pass which predicts the model outputs. This setup resembles the differentiable scheduled sampling proposed by \citet{goyal2017differentiable}.
\end{itemize}

\section{Experiments}

\begin{table}[hbt!]
\centering
\begin{tabular}{lr}
\hline
Encoder model type & Transformer \\
Decoder model type & Transformer \\
\# Enc. \& dec. layers & 6 \\
Heads & 8 \\
Hidden layer size & 512 \\
Word embedding size & 512 \\
Batch size & 32 \\
Optimizer & Adam \\
Learning rate & 1.0 \\
Warmup steps & 20,000 \\
Maximum training steps & 300,000 \\
Validation steps & 10,000 \\
Position Encoding & True \\
Share Embeddings & True \\
Share Decoder Embeddings & True \\
Dropout & 0.2 (DE-EN) \\
Dropout & 0.1 (JA-EN) \\
\hline
\end{tabular}
\caption{Hyperparameters shared across models}
\label{table:hyperparameters}
\end{table}

\begin{table*}[hbt!]
    \centering
    \begin{tabular}{lrrrr}
       \textbf{Experiment} & \multicolumn{2}{c}{\textbf{DE$-$EN}} & \multicolumn{2}{c}{\textbf{JA$-$EN}} \\
       & \textbf{Dev} & \textbf{Test} & \textbf{Dev} & \textbf{Test} \\
        \hline
        Teacher Forcing Baseline & 35.05 & \textbf{29.62} & 18.00 & 19.46 \\
        \hline
        \textbf{No backprop} &  &  &  &  \\
        Argmax & 23.99 & 20.57 & 12.88 & 15.13 \\
        Top-k mix & 35.19 & 29.42 & \textbf{18.46} & 20.24 \\
        Softmax mix $\alpha=1$ & 35.07 & 29.32 & 17.98 & 20.03 \\
        Softmax mix $\alpha=10$ & 35.30 & 29.25 & 17.79 & 19.67 \\
        Gumbel Softmax mix $\alpha=1$ & \textbf{35.36} & 29.48 & 18.31 & 20.21 \\
        Gumbel Softmax mix $\alpha=10$ & 35.32 & 29.58 & 17.94 & \textbf{20.87} \\
        Sparsemax mix & 35.22 & 29.28 & 18.14 & 20.15 \\
        \hline
        \textbf{Backprop through model decisions} &  &  &  &  \\
        Softmax mix $\alpha=1$ & 33.25 & 27.60 & 15.67 & 17.93 \\
        Softmax mix $\alpha=10$ & 27.06 & 23.29 & 13.49 & 16.02 \\
        Gumbel Softmax mix $\alpha=1$ & 30.57 & 25.71 & 15.86 & 18.76 \\
        Gumbel Softmax mix $\alpha=10$ & 12.79 & 10.62 & 13.98 & 17.09 \\
        Sparsemax mix & 24.65 & 20.15 & 12.44 & 16.23 \\
        \hline
    \end{tabular}
    \caption{Experiments with scheduled sampling for Transformer. 
    The table shows BLEU score for the best checkpoint on BLEU, measured on the validation set. The first group of experiments do not have a backpropagation pass through the first decoder. The results from the second group are from model runs with backpropagation pass through the second as well as through the first decoder. 
    }
    \label{table:results}
\end{table*}

We report experiments with scheduled sampling for Transformers for the task of machine translation. We run the experiments on two language pairs:

\begin{itemize}
    \item IWSLT 2017 German$-$English (DE$-$EN, \citet{cettolo2017overview}).
    \item KFTT Japanese$-$English (JA$-$EN, \citet{neubig2011kyoto}).
\end{itemize}
We use byte pair encoding (BPE; \citep{sennrich2016neural}) with a joint segmentation with 32,000 merges for both language pairs.


Hyperparameters used across experiments are shown in Table~\ref{table:hyperparameters}. All models were implemented in a fork of OpenNMT-py \citep{2017opennmt}. We compare our model to a \textbf{teacher forcing baseline}, i.e. a standard transformer model, without scheduled sampling, with the hyperparameters given in Table~\ref{table:hyperparameters}. 
We did hyperparameter tuning by trying several different values for dropout 
and warmup steps, 
and choosing the best BLEU score on the validation set for the baseline model.


With the scheduled sampling method, the teacher forcing probability continuously decreases over the course of training according to a predefined function of the training steps. Among the decay strategies proposed for scheduled sampling, we found that linear decay is the one that works best for our data:
\begin{equation}
    t(i) = \max\{\epsilon, k - ci\},
    \label{eq:linear_decay}
\end{equation}
where $0 \leq \epsilon < 1$ is the minimum teacher forcing probability to be used in the model and $k$ and $c$ provide the offset and slope of the decay. This function determines the teacher forcing ratio $t$ for training step $i$, that is, the probability of doing teacher forcing at each position in the sequence.


The results from our experiments are shown In Table~\ref{table:results}. The scheduled sampling which uses only the highest-scored word predicted by the model does not have a very good performance. The models which use mixed embeddings (the top-k, softmax, Gumbel softmax or sparsemax) and only backpropagate through the second decoder pass, perform slightly better than the baseline on the validation set, and one of them is also slightly better on the test set. 
The differentiable scheduled sampling (when the model backpropagates through the first decoder) have much lower results. The performance of these models starts degrading too early, so we expect that using more training steps with teacher forcing at the beginning of the training would lead to better performance, so this setup still needs to be examined more carefully.


\section{Discussion and Future Work}

In this paper, we presented our approach to applying the scheduled sampling technique to Transformers. Because of the specifics of the decoding, applying scheduled sampling is not straightforward as it is for RNN and required some changes in the way the Transformer model is trained, by using a two-step decoding. We experimented with several schedules and mixing of the embeddings in the case where the model predictions were used. We tested the models for machine translation on two language pairs. 
The experimental results showed that our scheduled sampling strategy gave better results on the validation set for both language pairs compared to a teacher forcing baseline 
and, in one of the tested language pairs (JA$-$EN), there were slightly better results on the test set. 

One possible direction for future work is experimenting with more schedules. We noticed that when the schedule starts falling too fast, for example, with the exponential or inverse sigmoid decay, the performance of the model degrades too fast.
Therefore, we think it is worth exploring more schedules where the training does more pure teacher forcing at the beginning of the training and then decays more slowly, for example, inverse sigmoid decay which starts decreasing after more epochs.
We will also try the experiments on more language pairs.

Finally, we need to explore the poor performance on the differential scheduled sampling setup (with backpropagating through the two decoders). In this case, the performance of the model starts decreasing earlier and the reason for this needs to be examined carefully. We expect this setup to give better results after adjusting the decay schedule to allow more teacher forcing training before starting to use model predictions.




\section*{Acknowledgments}
This work was %
supported by the European Research Council (ERC StG DeepSPIN 758969),
and by the Funda\c{c}\~ao para a Ci\^encia e Tecnologia 
through contracts UID/EEA/50008/2019 and CMUPERI/TIC/0046/2014 (GoLocal). 
We would like to thank Gon\c{c}alo Correia and Ben Peters for their involvement on an earlier stage of this project.

\bibliography{main}
\bibliographystyle{acl_natbib}

\end{document}